\pdfoutput=1

\documentclass[11pt]{article}

\usepackage{EMNLP2022}

\usepackage{times}
\usepackage{latexsym}

\usepackage[T1]{fontenc}

\usepackage[utf8]{inputenc}

\usepackage{microtype}

\usepackage{inconsolata}

\usepackage{comment}
\usepackage{paperpkg}
\usepackage{enumitem}
\usepackage{tikz}
\newcommand*\circled[1]{\tikz[baseline=(char.base)]{
            \node[shape=circle,draw,inner sep=0.5pt] (char) {#1};}}

\usepackage{amsmath}
\usepackage{amssymb}
\usepackage{mathtools}
\usepackage{amsthm}

\usepackage[capitalize,noabbrev]{cleveref}

\usepackage{times}
\usepackage{latexsym}
\usepackage{enumitem}
\usepackage{cleveref}
\usepackage{color, soul}
\crefname{section}{§}{§§}
\Crefname{section}{§}{§§}

\usepackage{tabularx}
\usepackage{xurl}

\theoremstyle{plain}

\theoremstyle{definition}

\newtheorem*{remark}{Remark}

\usepackage[textsize=tiny,disable=true]{todonotes}

\title{Inducer-tuning: Connecting Prefix-tuning and Adapter-tuning}

\author{
Yifan Chen$^{1}$\thanks{~~Equal contribution. This work was performed while the first author was interning at Amazon Alexa AI.} \quad
{\bf Devamanyu Hazarika}$^{2}$\footnotemark[1] \quad
{\bf Mahdi Namazifar}$^{2}$ \\
{\bf Yang Liu}$^{2}$ \quad
{\bf Di Jin}$^{2}$\thanks{~~Correspondence to: Di Jin <djinamzn@amazon.com>} \quad
{\bf Dilek Hakkani-Tur}$^{2}$ \\
$^1$University of Illinois Urbana-Champaign \quad
$^2$Amazon Alexa AI
}

\begin{document}
\maketitle
\begin{abstract}
Prefix-tuning, or more generally continuous prompt tuning, has become an essential paradigm of parameter-efficient transfer learning. Using a large pre-trained language model (PLM), prefix-tuning can obtain strong performance by training only a small portion of parameters. In this paper, we propose to understand and further develop prefix-tuning through the kernel lens. Specifically, we make an analogy between \textit{prefixes} and \textit{inducing variables} in kernel methods and hypothesize that \textit{prefixes} serving as \textit{inducing variables} would improve their overall mechanism. From the kernel estimator perspective, we suggest a new variant of prefix-tuning---\textit{inducer-tuning}, which shares the exact mechanism as prefix-tuning while leveraging the residual form found in adapter-tuning. This mitigates the initialization issue in prefix-tuning. Through comprehensive empirical experiments on natural language understanding and generation tasks, we demonstrate that inducer-tuning can close the performance gap between prefix-tuning and fine-tuning.
\end{abstract}

\section{Introduction}

Transfer learning from large pre-trained language models (PLMs) has been the de-facto method to tackle downstream natural language processing (NLP) tasks with proven performance and scalability~\citep{peters2018deep}.
Among all the adaption techniques, fine-tuning~\citep{howard2018universal, kale-rastogi-2020-text} is predominant for PLMs and maintains the models' architecture while updating all the parameters within. Though powerful, fine-tuning is considered parameter-inefficient since it results in separate copies of model parameters for each task/client after training.

With the sizes of PLMs increasing to hundreds of millions~\citep{brown2020language} or even up to trillion~\citep{fedus2021switch}  parameters, 
the trend motivates a range of parameter-efficient adaptation techniques, 
including \textit{adapter-tuning} and \textit{prompting}, as promising lightweight alternatives to fine-tuning to reduce computational consumption and storage space.
Adapter-tuning inserts bottlenecked Multi-layer Perception (MLP) modules between the pre-trained layers of PLMs and tunes only these new parameters for task adaptation \citep{houlsby2019parameter, pfeiffer-etal-2020-mad}.
Prompting, instead, aims to adapt the general-purpose PLMs through prompts, whose effectiveness has been shown on a frozen GPT-3 model \citep{brown2020language}.

An implicit drawback of the prompt-based adaptation is the difficulty of searching for the proper prompt.
To avoid manually designing the prompts, \citet{autoprompt:emnlp20} propose a search algorithm to find the effective prompt over discrete space of vocabularies;
prefix-tuning~\citep{Li2021PrefixTuningOC} and other concurrent methods~\citep{Lester2021ThePO, liu2021gpt, DBLP:journals/corr/abs-2110-07602} further extend the discrete search to continuous prompts, attaining performance close to fine-tuning in some tasks.
Despite the effort, there is still a performance gap between ``prefix-tuning'' and ``fine-tuning'' in many tasks, 
especially when the model size is small~\citep{Lester2021ThePO, he2021towards}. 
In addition, the mechanism of prefix-tuning is still poorly understood and under-explored.
Prefix-tuning is also similar to adapter-tuning,
since they both insert additional modules into each transformer layer (classical prompt-based methods \citep{Lester2021ThePO, liu2021gpt} only add prompts to the embedding layer).

Scrutinizing the evolution of prompt-based methods, we can observe they have gradually deviated from the concept of ``prompts''.
Compared to the manually designed prompts, the discrete search usually results in counter-intuitive prompt tokens,
which vaguely match the topic but are not as sensible as the manual one;
for continuous prompt tuning, it even breaks the limit of the existing vocabulary. 
All these pieces imply that the mechanism behind prompt-based tuning might be more complicated than guiding the output through hint prompts.
To open the black box of ``prompts'', in this work, we propose to consider the prompts (either hard or soft) as ``inducing variables'' in kernel methods \citep{titsias2009variational}. This analogy is justified due to the close connection between attention modules in PLMs and kernel estimators \citep{choromanski2020rethinking, chen2021skyformer, tsai2019transformer}.
This kernel perspective explains the potential mechanism of prefix-tuning and motivates a new method, \textit{inducer-tuning}.
Specifically, inducer-tuning freezes all the original parameters in the PLMs as other prompt-based methods; 
when computing the attention output for a certain input token in each layer, inducer-tuning utilizes a point close to the query vector as the ``inducer''. This unique ``soft prompt'' eases the search for appropriate prompts and builds a new connection between ``prompting'' and ``adapter-tuning''.

In summary,  the contribution  of this work is three-fold:
\circled{1} We explain the underlying mechanism of prefix-tuning as the inducing variables in kernel learning.
\circled{2} We propose a new parameter-efficient adaptation technique, inducer-tuning, to further improve prefix-tuning.
\circled{3} Through comprehensive empirical studies, we verify our proposed method can close the gap between ``prefix-tuning'' and ``fine-tuning'' on relatively small PLMs, and provide a tighter lower bound on the potential of continuous prompt tuning.

\section{Related Work}

In this section, we briefly introduce the classical form of adapter-tuning and mainly focus on the different variants of prompting.

\textbf{Adapter-tuning}.
Compared to fine-tuning all the parameters in the PLMs, 
\citet{houlsby2019parameter}, \citet{pfeiffer-etal-2020-mad} propose to modulate the output of a transformer layer through inserting additional small-bottleneck MLP layers (adapters) \citep{houlsby2019parameter}\footnote{We ignored layer normalization and bias terms here for brevity.}:
\begin{align}
\label{eqn:adapter}
\text{Adapter}(\mtx h) = \mtx h + \text{ReLU}(\mtx h \mtx{W}_1) \mtx{W}_2,
\end{align}
where $\mtx h$ is the dimension-$d$ hidden state in the transformer and $\mtx{W}_1, \mtx{W}_2$ are $d$-by-$r$ and $r$-by-$d$ projection matrices.
Adapters have a residual form similar to skip connection,
while only $\mtx{W}_1, \mtx{W}_2$ will be trained,
greatly decreasing the size of tunable parameters.
Up to now, the adapter-based method has been widely used for multiple NLP tasks~\citep{stickland2019bert,pfeiffer-etal-2020-mad,wang2020k,pfeiffer2020unks,ustun-etal-2020-udapter,vidoni2020orthogonal,pfeiffer-etal-2021-adapterfusion,he2021effectiveness,xu2021retrieval,ruckle2020adapterdrop,karimi2021compacter},
and adapters are also intrinsically connected to many other parameter-efficient adaptation techniques, as detailed in \citet{he2021towards}.

\textbf{Prompting}.
Prompting prepends task-specific instructions to the task input and was originally demonstrated in~\citet{brown2020language}.
As manual prompts rely on trial and error,  \citet{jiang-etal-2020-know}, \citet{autoprompt:emnlp20} suggests search algorithms to specify the prompts among all the tokens in the vocabulary.
Prompt-tuning \citep{Lester2021ThePO} and P-tuning~\citep{liu2021gpt} remove the vocabulary restriction on prompts by using trainable ``soft prompts''. The prompts in the aforementioned methods are only inserted into the bottom embedding layer of PLMs, 
while Prefix-tuning \citep{Li2021PrefixTuningOC,DBLP:journals/corr/abs-2110-07602} adds soft prompts to all the transformer layers to further increase the capacity of prompting.

Though effective, proper initialization of the soft prompts remains challenging. To mitigate the issue, \citet{Li2021PrefixTuningOC} used an extra MLP to re-parameterize the prompts in each layer, thus adding more parameters that need training;
SPoT \citep{vu2021spot} suggests performing pre-training for soft prompts using a wide range of NLP tasks, which requires additional computational resources.
In contrast, though adapters have a similar expression form to prefix-tuning~\citep{he2021towards}, 
adapter-tuning only requires regular initialization.
We speculate that the residual form of adapters mitigates the initialization issue since the output of each layer in the new model would be centered around the output in the frozen PLMs,
and the residual form contributes to gradient back-propagation as in skip connection.
We rely on this intuition and utilize the above-mentioned advantages of adapters to guide the design of our proposed inducer-tuning.

\section{Preliminaries: Transformer Layers}
\label{sec:transformer}

Before discussing the mechanism of prompt-tuning, we introduce the structure of transformer layers and necessary notations in this section.

A general transformer-based PLM is mainly composed of $L$ stacked layers. Each layer contains a multi-headed self-attention and a fully connected feed-forward network (FFN) sub-layer, 
both followed by an ``Add \& Norm'' module~\citep{vaswani2017attention}.
\footnote{For simplicity, we omit the cross-attention module in transformer-based encoder-decoder models.} 
Hereon, we shall focus on the structure of the attention sub-layer since prefix-tuning directly works on this sub-layer. 

Passing a length-$n$ input sequence $\mtx{X} \in \mb R^{n \times N_h p}$ to an attention sub-layer (assuming $N_h$ heads and dimension size $p$ for each head),
we first perform linear transforms to the input $\mtx{X}$ and obtain the query matrix $(\mtx{Q})$, the key matrix $(\mtx{K})$, and the value matrix $(\mtx{V})$ as:
\begin{align}
\mtx{Q/K/V} &= \mtx{X} \mtx{W}_{[q/k/v]} + \mtx{1} \mtx{b}_{[q/k/v]}^T, \label{eqn:linear_transform}
\end{align}
where $\mtx{Q}, \mtx{K}, \mtx{V} \in \mb R^{n \times N_hp}$ are the query/ key/ value matrix; 
$\mtx{W}_{[q/k/v]} \in \mb R^{N_hp \times N_hp}$ are the weight matrices, 
and $\mtx{b}_{[q/k/v]} \in \mb R^{N_h p}$ are the bias terms in the corresponding transformations.
\footnote{To ease the notations we adopt the practical setting where $\mtx{X}, \mtx{Q}, \mtx{K}, \mtx{V}$ have the same shape.}

To increase the model capacity, the three components $\mtx{Q}, \mtx{K}, \mtx{V}$ are respectively divided into $N_h$ blocks, contributing to the attention output in each head of the multi-headed self-attention module.
For instance, we represent $\mtx{Q}$ as $\mtx{Q} = \left( \mtx{Q}^{(1)}, \cdots, \mtx{Q}^{(N_h)} \right)$, 
where each block $\mtx{Q}^{(h)} = \mtx{X} \mtx{W}_q^{(h)} + \mtx{1} (\mtx{b}_q^{(h)})^T$ is an $n$-by-$p$ matrix,
and $\mtx{W}_q^{(h)}, \mtx{b}_q^{(h)}$ are the corresponding parts in $\mtx{W}_q, \mtx{b}_q$. 
The attention output for the $h^{th}$ head is:
\begin{align}
\mtx{L}^{(h)} \mtx{V}^{(h)} &\defeq \text{softmax}(\mtx{Q}^{(h)} (\mtx{K}^{(h)})^T / \sqrt{p}) \mtx{V}^{(h)} \nonumber \\ &= (\mtx{D}^{(h)})^{-1} \mtx{M}^{(h)} \mtx{V}^{(h)},
\label{eqn:attn}
\end{align}
where $\mtx{M}^{(h)} \defeq \exp\left(\mtx{Q}^{(h)} (\mtx{K}^{(h)})^T / \sqrt{p}\right)$ 
and $\mtx{D}^{(h)}$ is a diagonal matrix in which $\mtx{D}^{(h)}_{ii}$ is the sum of the $i$-th row in $\mtx{M}^{(h)}$, serving as the normalization procedure in softmax.
The attention outputs in each head are then concatenated as $\mtx{L} \defeq (\mtx{L}^{(1)} \mtx{V}^{(1)}, \dots, \mtx{L}^{(N_h)} \mtx{V}^{(N_h)})$.

After concatenating the heads, there is a linear transform following the output 
\begin{align}
\label{eqn:wo}
\mtx{L} \mtx{W}_o + \mtx{1} \mtx{b}_o^T, 
\end{align}
where $\mtx{W}_o$ and $\mtx{b}_o$ are similarly sized as the other matrices in~\cref{eqn:linear_transform}. This is the overall output of the attention sub-layer, 
which we shall revisit in~\Cref{sec:extension}.

\section{Parameter-Efficient Inducer-Tuning}
\label{sec:method}

We describe the motivation and the mechanism of inducer-tuning in this section.
We first revisit the connection between self-attention and kernel estimators in \Cref{sec:attn-as-kernel},
which interprets attention from another perspective by considering query, key, and value matrices as three separate sets of vectors rather than the related representations of the same input sequence.
This special perspective motivates and justifies the inducer-tuning we propose in \Cref{sec:inducer}.

\subsection{Attention as Kernel Estimators}
\label{sec:attn-as-kernel}

Traditionally, attention operation (\Cref{eqn:attn}) is viewed as a transformation $g(\cdot)$ of the input sequence $\mtx{X}$. However, in prefix-tuning, parameters within PLMs are frozen, which implies that given the input $\mtx{X}$, the represenattions $\mtx{Q}, \mtx{K}, \text{and~} \mtx{V}$ are invariant.\footnote{While our discussion is for a single attention head, we omit the superscript $(h)$ for brevity.}
This observation allows us to re-interpret attention as a kernel estimator $f(\cdot)$ with $\mtx{Q}$ as its input. Specifically, we denote the $i$-th input vector $\mtx{X}_i$'s attention operation as $f(\mtx{Q}_i) \defeq g(\mtx{X}_i)$. This attention representation can be seen as modifying the input query vector $\mtx{Q}_i$ to $f(\mtx{Q}_i)$ via \textit{supporting points} $\{\mtx{K}_j\}_{j=1}^n$~\citep{choromanski2020rethinking, peng2020random, chen2021skyformer}, which can be considered as a Nadaraya–Watson kernel estimator \citep[Definition~5.39]{wasserman2006all}:
\begin{align*}
\text{row-normalize}\left(\kappa\left(\mtx Q, \mtx K\right) \right) \mtx V,
\end{align*}
where $\kappa(\cdot, \cdot)$ is a kernel function.
(Refer to~\Cref{sec:kernel_estimator} for more details on this claim.)

\subsection{Prefix-Tuning and Inducing Variables}
\label{sec:prefix}

Prefix-tuning \citep{Li2021PrefixTuningOC} alters the attention output in each layer.
Concretely, it prepends length-$l$ prefix vectors $\mtx{P}_k, \mtx{P}_v \in \mb R^{l \times p}$ to $\mtx{K}$ and $\mtx{V}$, respectively; 
for a certain query token $\mtx{Q}_i$ (the $i$-th row of the query matrix $\mtx{Q}$), 
its attention output $f(\mtx{Q}_i) \defeq \text{Attn}(\mtx{Q}_i, \mtx{K}, \mtx{V})$
is updated as a weighted sum of $f(\mtx{Q}_i)$ and $\text{Attn}(\mtx{Q}_i, \mtx{P}_k, \mtx{P}_v)$ \citep[Equation~(7)]{he2021towards}.

\begin{remark}
From the kernel estimator perspective, the two categories of virtual tokens play different roles.
The virtual key vectors $\mtx{P}_k$ apply to the empirical kernel matrix part and can alter the attention scores (and thus the weights for $\text{Attn}(\mtx{Q}_i, \mtx{P}_k, \mtx{P}_v)$);
whereas $\mtx{P}_v$ takes effect in the value part.
It might not be optimal for prefix-tuning to model the two categories of virtual tokens similarly. 
In \Cref{sec:inducer} we will show how \textit{inducer-tuning} addresses the two parts through different residual forms.
\end{remark}

We suggest that the mechanism of prefix-tuning can be further understood through the concept of \textit{inducing variables} in kernel learning literature \citep{titsias2009variational}.
Many computational methods in kernel learning utilize a small set of support points (inducing variables) to improve the inference performance \citep{Musco2017RecursiveSF, chen2021fast}.
\citet{Snelson2005SparseGP} specifically consider the inducing variables as auxiliary pseudo-inputs and infer them using continuous optimization, 
which is similar to prefix-tuning.
We emphasize that from the first sight the main character of inducing-point methods is representing a vast amount of training examples through a small number of points, so as to reduce the computational cost;
however, here we instead aim to leverage the mechanism of inducing variables to  well-steer the estimation: the goal we try to attain is to strengthen prefix-tuning by making the prefixes better modulate the attention output.
We introduce and analyze the mechanism as follows.

\textbf{Mechanism for well-steering inference outputs in inducing-point methods}. 
Conceptually, inducing variables help the inference because they can represent the distribution of the query inputs and steer the kernel methods without changing the kernel in use.
In particular, we consider the distribution pattern of unconstrained inducing points $X_M$~\citep[Figure~1]{Snelson2005SparseGP}. 
We observe that most of them are \textit{close to the testing examples} $X^*$,
and in the new estimation~\citep[Equation~(8)]{Snelson2005SparseGP} the inducers $X_M$ will receive great weights through the weights assignment mechanism in kernel methods (we recall kernel methods can assign the weights of samples as attention \citep{choromanski2020rethinking, chen2021skyformer, tsai2019transformer}; for inducing variables close to the query, they would automatically receive more attention), and thus effectively modulate the output. 

From this mechanism, we draw an inductive bias "the prefix should be close to the query" (which is not enforced in the method of prefix-tuning) and accordingly propose inducer-tuning.
We remark since we are not pursuing the original goal, reducing computational cost, of inducing variables, it is ordinary that the concrete design in the next subsection is different from the usual form of inducing points, a small number of samples.

We speculate prefix-tuning partially benefits from the above mechanism as well.
Furthermore, some indirect evidence is stated as follows.
As discussed in previous studies, to make the full potential of prompting, the manually designed prompts are expected to be related to the topic of the input sequence \citep{brown2020language} (close to the query);
even for the soft prompts they are recommended to be initialized with the token relevant to the specific tasks \citep{Li2021PrefixTuningOC},
which also requires the prompts to be close to the query to provide effective adaptation.
With this belief, we propose \textit{inducer-tuning} to exploit further the mechanism of inducing variables and improve upon prefix-tuning.

\subsection{Method}
\label{sec:inducer}

Inducer-tuning follows the same design principle as prefix-tuning, which modulates the attention output through inserting virtual tokens (vectors). However, unlike prefix-tuning, our virtual tokens are not shared among the input sequences.
Inducer-tuning also incorporates the benefits of residual forms to ease the initialization and remove the re-parametrization trick in prefix-tuning.
Specifically, we suggest the following modifications:
\circled{1} The ``inducers'' are adaptive to and customized for each input token to strengthen the expressiveness of the new attention output.
\circled{2} We propose to model the virtual vectors in a residual form as an adapter, which makes the final attention output be in a residual form as well.
We now dive into discussing the intuitions behind the modifications in detail.

\textbf{Adaptive \textit{inducers}.}
There is an important difference between language models and kernel methods,
making fixed prefixes less effective than inducing variables in kernel methods.
In language models, the distribution of the input queries keeps changing, 
and for some inputs, the fixed prefixes fail to be qualified as ``inducing variables''.
Even worse, for a long input, there probably exists some query vectors away (regarding $\ell_2$ distance) from all the virtual vectors in the fixed prefixes,
which are thus unable to modulate the attention output well.
The phenomenon that prefix-tuning has a relatively poorer performance on tasks with longer inputs can be observed in our experiments (\Cref{sec:results}).

To alleviate the above issue, we propose adaptive modeling of the virtual key vectors.
For a query $\mtx{Q}_i$, we suggest taking a vector close to $\mtx{Q}_i$ itself as the corresponding virtual key vector (the length of the new prefix is thus $1$), in the hope of leading to better inference.

As for the virtual value vectors,
we relate them to the corresponding virtual key vectors.
The motivation comes from traditional (non-self-)attention, 
whose mechanism coincides with a kernel estimator: 
the value $\mtx{V}$ is independent of the query sequence $\mtx{Q}$ and related to the supporting points $\mtx{K}$.

Specifically, considering our design above that the virtual key vectors are close to $\mtx{Q}_i$ (we take the virtual key vectors as transforms of the input query vectors $\mtx{Q}_i$'s), we propose to accordingly model the virtual value vectors as a map of $\mtx{Q}_i$ as well,
which implies the virtual value vectors are also adaptive to the input query vectors.

\paragraph{Adapter Structures.}
To stabilize the training procedure, we propose incorporating the adapter structures into modeling the virtual key/value vectors.
Specifically, for the $i$-th token $\mtx{Q}_i$ (in a certain head), we represent the corresponding virtual key/value vectors respectively as
\begin{align}
\mtx{P}_{k, i} &= \mtx{Q}_i + \text{MLP}_k(\mtx{Q}_i) \\
\mtx{P}_{v, i} &= f(\mtx{Q}_i) + \text{MLP}_v(\mtx{Q}_i),
\label{eqn:Pvi}
\end{align}
where $\text{MLP}_{k/v}$ will both return a vector of the same dimension as the input $\mtx{Q}_i$.
\footnote{Note that these virtual vectors can be applied to causal attention in auto-regressive decoders since they do not utilize any future token information.}

It is natural to model $\mtx{P}_{k, i}$ in a residual form as in Equation~(\ref{eqn:adapter}), 
considering $\mtx{P}_{k, i}$ is expected to center around $\mtx{Q}_i$;
as for $\mtx{P}_{v, i}$, we claim the specific form in Equation~(\ref{eqn:Pvi}) allows the complete expression of inducer-tuning to be adapter-like,
and the justification is stated as the following derivation.

\begin{figure*}[t]
    \centering
    \includegraphics[width=\textwidth]{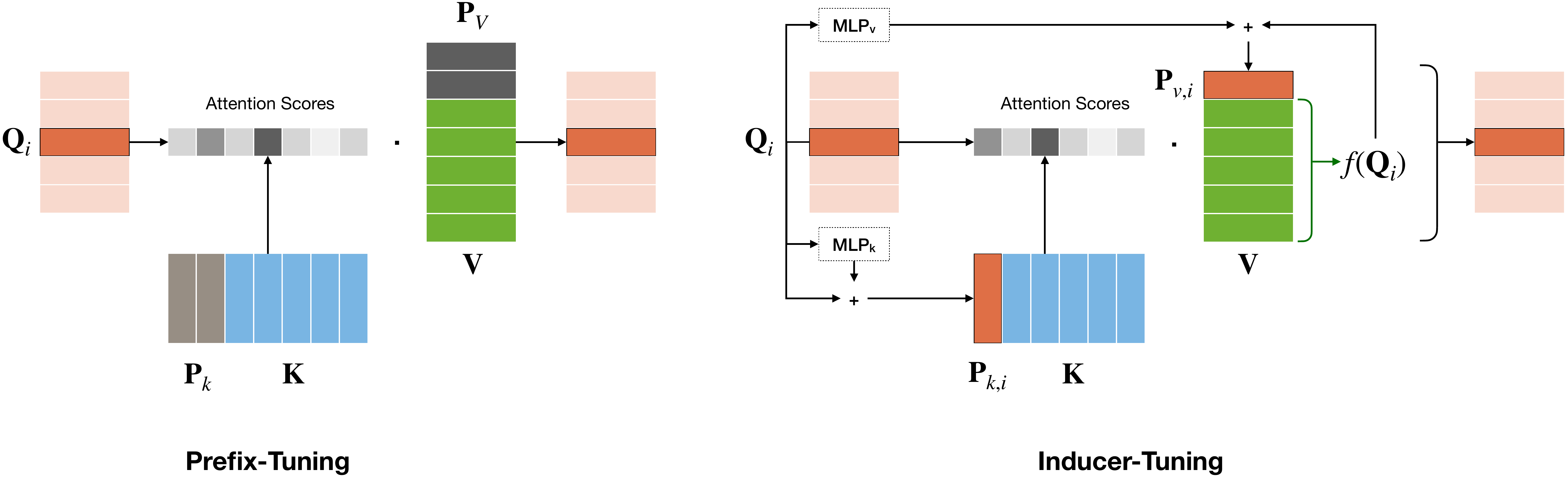}
    \caption{The mechanisms of prefix-tuning (left) and inducer-tuning (right) in inference (the MLP module for reparameterization in prefix-tuning is dropped). 
    For prefix-tuning, the virtual tokens ($\mtx{P}_k, \mtx{P}_v$) are shared among all the query vectors;
    inducer-tuning instead prepends customized inducers ($\mtx{P}_{k, i}, \mtx{P}_{v, i}$) for a certain vector $\mtx{Q}_i$.}
    \label{fig:my_label}
\end{figure*}

To derive the expression for inducer-tuning, 
we denote the new key matrix and value matrix (specific to the input query vector $\mtx{Q}_i$) as 
\begin{align*}
\mtx{\wt{K}}^{(i)} = 
\begin{bmatrix}
    \mtx{P}_{k, i}^T \\
    \mtx{K}
\end{bmatrix}, \quad
\mtx{\wt{V}}^{(i)} = 
\begin{bmatrix}
    \mtx{P}_{v, i}^T \\
    \mtx{V}^T
\end{bmatrix}.
\end{align*}
The new attention output $\tilde f(\mtx{Q}_i)$ for the query $\mtx{Q}_i$ is thus (omitting the factor $1/\sqrt{p}$ for clarity)
\begin{align}
& \text{Attn}(\mtx{Q}_i, \mtx{\wt{K}}^{(i)}, \mtx{\wt{V}}^{(i)}) \nonumber \\ 
=& \frac{\exp(\dotp{\mtx{Q}_i}{\mtx{P}_{k, i}}) \mtx{P}_{v, i} + \sum_j \exp(\dotp{\mtx{Q}_i}{\mtx{K}_j}) \mtx{V}_j}
{\exp(\dotp{\mtx{Q}_i}{\mtx{P}_{k, i}}) + \sum_j \exp(\dotp{\mtx{Q}_i}{\mtx{K}_j})} \nonumber \\
=& \lambda_i \mtx{P}_{v, i} + (1 - \lambda_i) f(\mtx{Q}_i)
\label{eqn:weighted_sum}
\end{align}
where we define the weight $\lambda_i$ as,
\begin{align*}
\frac{\exp(\dotp{\mtx{Q}_i}{\mtx{P}_{k, i}})}{\exp(\dotp{\mtx{Q}_i}{\mtx{P}_{k, i}}) + \sum_j \exp(\dotp{\mtx{Q}_i}{\mtx{K}_j})}.
\end{align*}

Combining the pieces, we state the complete equation for the new attention output $\tilde f(\mtx{Q}_i)$ as,
\begin{align}
& \lambda_i \mtx{P}_{v, i} + (1 - \lambda_i) f(\mtx{Q}_i) \nonumber \\
=& \lambda_i \left( f(\mtx{Q}_i) + \text{MLP}_v(\mtx{Q}_i) \right) + (1 - \lambda_i) f(\mtx{Q}_i) \nonumber \\
=& f(\mtx{Q}_i) + \lambda_i \text{MLP}_v(\mtx{Q}_i).
\label{eqn:simplified}
\end{align}
We observe inducer-tuning now perturbs the output $f(\mtx{Q}_i)$ in a residual form,
which therefore \textit{connects} prefix-tuning and adapter-tuning.

The procedure of inducer-tuning is summarized in Figure~\ref{fig:my_label},
and \Cref{sec:ablation} shows the residual form greatly impacts the model performance.

\subsection{Extending the Scope of Value}
\label{sec:extension}

Besides the representation of the virtual vectors,
we propose another improvement via the self-attention decomposition proposed by \citet{hou2020dynabert}.

Considering the linear transform right after the attention module, 
we can accordingly rewrite the attention sub-layer as (ignoring the bias term in the linear transform)
\begin{align*}
\sum_{h=1}^{N_h} \text{softmax}(\mtx{Q}^{(h)} (\mtx{K}^{(h)})^T / \sqrt{p}) \mtx{V}^{(h)} \mtx{W}_o^{(h)},
\end{align*}
where $\mtx{W}_o^{(h)}$ is the $h$-th row block in $\mtx{W}_o$.
Notably, $\mtx{W}_o^{(h)}$ is attached to the value matrix $\mtx{V}^{(h)}$, 
suggesting that $\mtx{W}_o^{(h)}$'s should be counted into the complete kernel structure of a head.
We therefore define the complete attention output $\bar f(\mtx{Q}^{(h)})$ as 
\begin{align}
\text{softmax}\left({\mtx{Q}^{(h)} (\mtx{K}^{(h)})^T}/{\sqrt{p}} \right) \mtx{V}^{(h)} \mtx{W}_o^{(h)},
\label{eqn:complete_head}
\end{align}
and align the prefix vectors $\mtx{P}_{v, i}$'s with the rows in $\mtx{V}^{(h)} \mtx{W}_o^{(h)}$, instead of solely $\mtx{V}^{(h)}$ as in prefix-tuning.
The detailed implementation of the extended $\mtx{P}_{v, i}$ is provided in Appendix~\ref{sec:implementation}.
We can verify the improvement by this extension through the ablation studies in \Cref{sec:ablation}.

\subsection{A Potential Limitation of Prompting}
\label{sec:limitation}

A potential limitation of prompt-based methods comes from the frozen weight matrices $\mtx{W}_q$ and $\mtx{W}_k$.
For all the $n(n+l)$ pairs of query / key vectors in a head, 
most of the pairs (corresponding to the elements within $\mtx{Q} \mtx{K}^T$) have invariant pairwise positional interactions due to the frozen weight matrices $\mtx{W}_q$ and $\mtx{W}_k$.
However, on downstream tasks, there can be a mismatch between $\mtx{W}_q$ and $\mtx{W}_k$ maintained from pre-training:
the distribution of $\mtx{Q}, \mtx{K}$ will substantially change due to the distinct task-specific datasets as well as the virtual tokens added in the previous layers.
There is no adaptation to ensure the positional interactions between $\mtx{Q}, \mtx{K}$ still contribute to the proper representation $f(\mtx{Q}_i)$.

To resolve the potential issue of prefix-tuning,
we suggest applying low-rank adaptation (LoRA)~\citep{hu2021lora} to $\mtx{W}_q$ as a complement to prompt-based methods, including inducer-tuning.
Specifically, before we compute the attention output in each layer,
$\mtx{W}_q$ will be updated as
\begin{align}
\mtx{W}_q \leftarrow \mtx{W}_q + \mtx{B} \mtx{A},
\label{eqn:lora}
\end{align}
where $\mtx{W}_q$ is kept frozen and $\mtx{B} \in \mb R^{N_h p \times r}, \mtx{A} \in \mb R^{r \times N_h p}$ will be tunable in training.
We report in~\Cref{sec:results} that combining both inducer-tuning and LoRA outperforms their individual counterparts.

\textbf{Final Model.}
Our final proposed model does the inferencex as follows:
\circled{1} in each layer, we first apply Equation~(\ref{eqn:lora}) to update $\mtx{W}_q$ before obtaining $\mtx{Q}, \mtx{K}, \mtx{V}$;
\circled{2} construct the inducer matrices $\mtx{P}_k = \mtx{Q} + \text{MLP}_k(\mtx{Q})$,
and compute the vector $\mtx{a}$ with the $i$-th component $\mtx{a}_i = \dotp{\mtx{Q}_i}{\mtx{P}_{k, i}}$;
\circled{3} compute the matrix product $[\mtx{a}; \mtx{Q} \mtx{K}^T]/\sqrt{p}$ and then perform softmax over the product---the first column (denoted as $\mtx{p}$) is the weights $\lambda_i$'s in Equation~(\ref{eqn:weighted_sum});
\circled{4} obtain $\bar f(\mtx{Q})$ as in Equation~(\ref{eqn:complete_head}), 
and return $\bar f(\mtx{Q}) + \text{diag}(\mtx{p}) \overline{\text{MLP}}_v(\mtx{Q})$ (corresponding to Equation~(\ref{eqn:simplified})) as the complete attention output.

\begin{table*}[h!]
    \centering
    \vskip 0.15in
    \resizebox{\textwidth}{!}{
    \begin{tabular}{|lcc|ccccccccc|cc|}
        \hline
        &&& \multicolumn{9}{c}{\textbf{WebNLG}} & \multicolumn{2}{|c|}{\textbf{CoQA}} \\
        &\multicolumn{2}{c|}{Parameters to}& \multicolumn{3}{c}{BLEU}  & \multicolumn{3}{c}{MET} & \multicolumn{3}{c|}{TER $\downarrow$}  & EM & F1 \\
        & train & store  & S & U & A & S & U & A & S & U & A & \multicolumn{2}{c|}{ } \\
        \hline
        Fine-tuning       & 100.00\% & 100.00\% & 59.8          & 28.7          & 46.1          & 0.43          & 0.29          & 0.36          & 0.38          & 0.68          & 0.51          & 59.0          & 67.4          \\
        \hline
        Adapter-108       & 1.62\%   & 1.62\%   & 59.5          & 34.1          & 48.2          & 0.42          & 0.32          & \textbf{0.38} & 0.38          & 0.61          & 0.49          & 57.7          & 66.4                    \\
        LoRA-54           & 1.61\%   & 1.61\%   & 54.8          & 36.9          & 46.7          & 0.40          & {0.33} & 0.37          & 0.41          & \textbf{0.55} & 0.47          & 57.2          & 65.7                    \\
        Prefix-tuning-108 & 7.98\%   & 1.60\%   & 56.1          & {37.2} & 47.6          & 0.40          & {0.33} & 0.37          & 0.40          & \textbf{0.55} & 0.47          & 51.8          & 60.3                    \\
        MAM-adapter    & 1.61\%   & 1.61\%   & 58.9 & 36.2 & 48.7 & 0.42 & 0.33 & 0.38 & 0.38 & 0.59 & 0.47          & 56.4          & 65.0                    \\
        \hline
        Inducer-tuning    & 1.61\%   & 1.61\%   & 59.4 & 36.8 & 49.2 & 0.42 & 0.33 & 0.38 & 0.38 & 0.59 & 0.47          & 57.7          & 66.1                    \\
        $\quad$ + LoRA     & 1.61\%   & 1.61\%   & 59.8 & \textbf{37.5} & \textbf{49.7}* & 0.43 & \textbf{0.34}* & \textbf{0.38} & \textbf{0.37}* & 0.57 & \textbf{0.46}*         & 58.7          & 67.1  \\
        MAM inducer-tuning    & 1.61\%   & 1.61\%   & \textbf{59.9}* & 36.8 & 49.5 & \textbf{0.43} & 0.33 & \textbf{0.38} & \textbf{0.37}* & 0.57 & 0.47          & \textbf{59.9}*  & \textbf{68.4}*                    \\
        \hline
        \end{tabular}
    }
    \caption[Caption of table]{\label{tab:result} Performance (\%) on WebNLG \textsuperscript{a} and CoQA. 
    All the parameter-efficient methods have similar sizes of parameters to store.
    The best scores (among parameter-efficient methods) under different metrics are \textbf{boldfaced} (for TER, the lower the metric is, the better the performance is). 
    Significance tests are performed between our methods and the other baselines for each metric ($5$ runs for WebNLG and $3$ runs for the others),
    and a superscript * is added if the test p-value $<0.05$.
    }
    \vskip 0.15in
    \raggedright
    \small\textsuperscript{a} For the metrics on WebNLG, the notations are same as in prefix-tuning \citep[Table~1]{Li2021PrefixTuningOC} that S, U, and A denote SEEN, UNSEEN, and ALL respectively; 
    in training only the examples from the \underline{S}EEN categories are used; 
    the examples from the \underline{U}NSEEN categories only appear in the test set; 
    and \underline{A}LL consists of all the categories.
\end{table*}

\begin{table}[!ht]
    \centering
    \resizebox{\columnwidth}{!}{
    \addtolength{\tabcolsep}{-2pt} 
    \begin{tabular}{|l|c|c|}
\hline 
Method (\# params) & MNLI & SST2 \\
\hline 
Fine-tuning (100\%) & $87.6_{\pm .4}$ & $94.6_{\pm .4}$ \\
\hline 
Bitfit $(0.1 \%)$ & $84.7$ & $93.7$ \\
Prefix-tuning $(0.5 \%)$ & $86.3_{\pm .4}$ & $94.0_{\pm .1}$ \\
LoRA $(0.5 \%)$ & $87.2_{\pm .4}$ & $94.2_{\pm .2}$ \\
Adapter $(0.5 \%)$ & $87.2_{\pm .2}$ & $94.2_{\pm .1}$ \\
MAM-Adapter $(0.5 \%)$ & $\textbf{87.4}_{\pm .3}$ & $94.2_{\pm .3}$ \\
\hline
Inducer-tuning $(0.5 \%)$ & $86.6_{\pm .2}$ & $94.1_{\pm .3}$ \\
Inducer-tuning + LoRA  $(0.5 \%)$ & $86.8_{\pm .5}$ & $94.7_{\pm .3}$ \\
MAM inducer-tuning $(0.5 \%)$ & $\textbf{87.4}_{\pm .04}$ & $\textbf{94.8}_{\pm .3}$ \\
\hline
\end{tabular}
}
\caption{\label{tab:nlu}
Accuracy (\%) on MNLI and SST2.
Baseline numbers and settings are from \citet{he2021towards}.}
\end{table}

\section{Experiments}

While prefix-tuning has been shown comparable to fine-tuning on some natural language understanding (NLU) tasks \citep{DBLP:journals/corr/abs-2110-07602}, 
there is still a performance gap between prefix-tuning and fine-tuning on natural language generation (NLG) tasks, especially for those tasks with long input sequences. 
Complete settings of the experiments below can be found in Appendix~\ref{sec:appendix_dataset} and Appendix~\ref{sec:appendix_training}.
The code for our algorithms is publicly available at \url{https://github.com/ychen-stat-ml/kernel-adapters}.

\subsection{Sketch of the Tasks}
\label{sec:task}

We test the performance of our methods on both NLU and NLG tasks.
For NLU tasks, we follow \citep{he2021towards} to use $\text{RoBERTa}_{\text{BASE}}$ \citep{liu2019roberta} on MNLI~\citep{N18-1101} and SST2~\citep{socher2013recursive} from the GLUE benchmark~\citep{wang2019glue}; 
in SST2, the models predict the two-way sentiment (positive/negative) of a given sentence,
and the MNLI task is to decide, given a premise and a hypothesis, 
whether there is entailment, contradiction, or neither.
We use $\text{GPT-2}_{\text{SMALL}}$ \citep{radford2019language} for NLG tasks:
WebNLG-challenge \citep{gardent2017creating} focuses on table-to-text tasks, 
in which the language models generate some relatively long and sensible sentences based on the triples with solely a few words;
in contrast, CoQA \citep{reddy2019coqa} provides the data for conversational question answering
\footnote{The official validation set is taken as the test set in our experiments, while we randomly choose $500$ instances from the training set as the new validation set.}, 
which requires the language model to return short answers to questions based on long conversational materials.
More details about the datasets (including the average sequence length) and the evaluation metrics used are provided in Appendix~\ref{sec:appendix_dataset}.

\subsection{Baselines}
\label{sec:candidate}

We compare our method with other representative methods:
Fine-Tuning \citep{howard2018universal}, Adapters \citep{houlsby2019parameter} used by \citet{lin2020exploring}, Prefix-Tuning \citep{Li2021PrefixTuningOC}, and LoRA~\citep{hu2021lora}
\footnote{LoRA in our experiments follows the recommended setting by \citet{hu2021lora}, adjusting both $\mtx{W}_q$ and $\mtx{W}_v$.};
we also follow the strategy in Mix-And-Match (MAM) adapters~\citep{he2021towards} to combine inducer-tuning (in self-attention) with adapters in FFN sub-layers,
and study how the combination compares to the original MAM adapter (prefix-tuning $+$ adapters in FFN).
In Table~\ref{tab:result} the suffixes after adapters / prefix-tuning / LoRA indicate the corresponding bottleneck size / prefix length / rank of updates, respectively. 

We differentiate the number of parameters to store and tune, 
as for prefix-tuning, the two numbers are inconsistent due to a re-parametrization trick \citep{Li2021PrefixTuningOC} to mitigate the initialization issue.
Instead of directly setting up an embedding matrix for virtual tokens, an additional MLP module in each layer is used in prefix-tuning to model the representation for those virtual tokens;
after the fine-tuning stage, the additional MLP modules are dropped and only the output embedding for virtual tokens needs storing, which leads to a regular number of parameters to store.
For the proposed inducer-tuning, we adopt the residual form to address the initialization issue and avoid the usage of the extra MLP, which makes inducer-tuning have the same number of parameters to store as to train and behave more like a regular adapter.

To make a fair comparison, we intentionally choose the number of parameters to \textbf{store} in prefix-tuning roughly the same as its adapter counterpart by adjusting the prefix length.
Detailed settings are available in~\Cref{sec:method_setting}.

\begin{table*}[h!]
    \centering
    \vskip 0.15in
    \resizebox{\textwidth}{!}{
    \begin{tabular}{|lcc|ccccccccc|cc|}
        \hline
        &&& \multicolumn{9}{c}{\textbf{WebNLG}} & \multicolumn{2}{|c|}{\textbf{CoQA}} \\
        &\multicolumn{2}{c|}{Parameters to}& \multicolumn{3}{c}{BLEU}  & \multicolumn{3}{c}{MET} & \multicolumn{3}{c|}{TER $\downarrow$}  & EM & F1 \\
        & train & store  & S & U & A & S & U & A & S & U & A & \multicolumn{2}{c|}{ } \\
        \hline
Prefix-tuning-108 & 7.98\% & 1.60\% & 56.1 & 37.2 & 47.6 & 0.40 & 0.33 & 0.37 & 0.40 & 0.55 & 0.47 & 51.8 & 60.3 \\
\hline
Adaptive          & 1.62\% & 1.62\% & 57.5 & 36.9 & 48.2 & 0.41 & 0.33 & 0.37 & 0.39 & 0.57 & 0.47 & 55.4 & 63.9 \\
Extension         & 1.55\% & 1.55\% & 59.1 & 36.7 & 49.0 & 0.42 & 0.33 & 0.38 & 0.38 & 0.58 & 0.47 & 57.2 & 65.7 \\
Gating            & 1.61\% & 1.61\% & 57.1 & 29.2 & 44.6 & 0.41 & 0.286 & 0.35 & 0.40 & 0.65 & 0.51 & 49.1 & 58.5 \\
\hline
Inducer-tuning    & 1.61\% & 1.61\% & 59.4 & 36.8 & 49.2 & 0.42 & 0.33 & 0.38 & 0.38 & 0.59 & 0.47 & 57.7 & 66.1 \\
$\quad$ + LoRA    & 1.61\% & 1.61\% & 59.8 & 37.5 & 49.7 & 0.43 & 0.34 & 0.38 & 0.37 & 0.57 & 0.46 & 58.7 & 67.1 \\
\hline
\end{tabular}
}
\caption{\label{tab:ablation} 
    Compare several variants of inducer-tuning against Prefix-tuning-108 and our proposed methods (copied from Table~\ref{tab:result}). 
    The exact settings of the methods listed are illustrated in \Cref{sec:ablation} and Appendix~\ref{sec:method_setting}.
    }
\end{table*}

\section{Results}
\label{sec:results}

\subsection{Main Results}

We conclude our experimental results in Tables~\ref{tab:result}~and~\ref{tab:nlu}, 
comparing the proposed inducer-tuning (\Cref{sec:inducer}), or inducer-tuning with LoRA (\Cref{sec:limitation}), against other baselines.
The benefit of using Mix-And-Match (MAM) techniques \citep{he2021towards} is also explored and stated as follows.

\paragraph{Performance of our proposed methods.}
Our proposed methods generally improve the performance of prompt-based methods.
The average accuracy of MAM inducer-tuning on MNLI and SST2 is increased by 0.8\% compared to Prefix-tuning.
The benefits are clearer on NLG tasks:
on WebNLG, inducer-tuning with LoRA provides a 1.5\% increase in BLEU score compared to Adapter-108 and a 2\%+ increase compared to LoRA-54 and Prefix-tuning-108;
on CoQA, all the previous parameter-efficient methods cannot attain a close performance to fine-tuning on this harder task,
while inducer-tuning with LoRA closes this gap, which shrinks to 0.3\% with solely 1.61\% tunable parameters of $\text{GPT-2}$.

\paragraph{The MAM technique benefits inducer-tuning.}
As remarked by \citet{he2021towards}, the ``Mix-And-Match'' of adapters in both self-attention and FFN sub-layers can better exploit parameter-efficient transfer learning than only modulating a single sub-layer.
We obtain a similar conclusion by replacing prefix-tuning with inducer-tuning ($+$ LoRA) in self-attention sub-layers.
The combination (MAM inducer-tuning) performs well on most of the tasks;
especially on the tasks with relatively longer sequences, MNLI and CoQA, 
MAM inducer-tuning attains respectively 0.6\% and 1.2\% performance improvement over vanilla inducer-tuning $+$ LoRA.

\paragraph{Long inputs deteriorate prefix-tuning.}
Notably, the performance of prefix-tuning is sensitive to the input length (c.f. \Cref{sec:inducer}).
For WebNLG with short inputs, prefix-tuning attains comparable performance with fine-tuning and other parameter-efficient methods.
On CoQA, however, prefix-tuning has a substantially lower exact-match / F1 score than others (e.g., over 7\% decrease in F1 score compared with fine-tuning).
{The similar pattern can be observed on the two NLU tasks as well: 
the performance gap between prefix-tuning and other candidate methods is much smaller on SST2, 
whose mean sequence length is shorter than MNLI.}
We remark our proposed adaptive inducers somewhat resolve the issue: 
both variants of inducer-tuning in Table~\ref{tab:result} obtain a 5\%+ improvement on CoQA. 

\paragraph{Enhance inducer-tuning through adapting pairwise positional interactions.}
In \Cref{sec:limitation}, we speculate the prompt-based methods can benefit from adapting pairwise positional interactions, 
and we investigate it on both NLU and NLG tasks. 
With the same parameter budgets, the inducer-tuning $+$ LoRA outperforms the pure inducer-tuning on all tasks.
The improvement is more evident in CoQA, the more challenging generation task with longer input sequences.
We remark that inducer-tuning more effectively exploits the tunable parameters than LoRA-54 for the value part,
as the combination variant also performs better than pure LoRA.

\subsection{Ablation Studies}
\label{sec:ablation}

We perform ablation studies on generation tasks to analyze the efficacy of the different components in our proposed method.
We recall there are four different features in inducer-tuning compared to prefix-tuning,
including the usage of adaptive inducers, the extension of virtual value vectors, the residual form of $\mtx{P}_k$, and the design for $\mtx{P}_{v, i}$ to concentrate around attention output.

Accordingly, we implement three other variants of inducer-tuning to help ablate the effects of the above-mentioned components.
Among them, \textit{Adaptive} directly takes $\mtx{Q}_i$ as $\mtx{P}_{k, i}$ but still models $\mtx{P}_{v, i}$ as $f(\mtx{Q}_i) + \text{MLP}_v(\mtx{Q}_i)$;
upon Adaptive, \textit{Extension} changes $\mtx{P}_{v, i}$ to $\bar f(\mtx{Q}_i) + \overline{\text{MLP}}_v(\mtx{Q}_i)$;
compared to Extension, \textit{Inducer-tuning} just modifies $\mtx{P}_{k, i}$ to $\mtx{Q}_i + {\text{MLP}}_k(\mtx{Q}_i)$;
to justify the design that $\mtx{P}_{v, i}$ centers around the attention output, \textit{Gating} models $\mtx{P}_{v, i}$ simply as $\overline{\text{MLP}}_v(\mtx{Q}_i)$, and the new complete attention output thus becomes $(1-\lambda_i)\bar{f}(\mtx{Q}_i) + \lambda_i \overline{\text{MLP}}_v(\mtx{Q}_i)$.
The concrete setting of each variant is deferred to Appendix~\ref{sec:method_setting} due to limited space.

\textbf{The usage of adaptive inducers}.
To demonstrate the benefits of adaptive inducers, we compare Prefix-tuning-108 with the basic counterpart---Adaptive.
Table~\ref{tab:ablation} shows Adaptive attains close performance to Prefix-tuning-108 on WebNLG
while obtaining a substantial improvement on CoQA, which has longer inputs.

\textbf{The extension of virtual value vectors}.
We observe an obvious improvement attributed to extending the scope of virtual value vectors by comparing the performance of Adaptive and Extension.
For almost all the metrics, Extension obtains better performance than Adaptive, with the same number of tunable parameters.

\textbf{The residual form of $\mtx{P}_k$}.
A natural design for $\mtx{P}_k$ is to directly model it as $\mtx{Q}$, which would automatically be the closest vectors to the ones in $\mtx{Q}$.
To ablate the usage of $\text{MLP}_k$, we compare Inducer-tuning against Extension, which follows the natural design to model $\mtx{P}_k$.
Through the empirical results, we find assigning parameters to $\text{MLP}_k$ can still slightly help the performance of inducer-tuning.

\textbf{$\mtx{P}_{v, i}$ centers around $\bar f(\mtx{Q}_i)$}.
Lastly, to show the benefits of modeling $\mtx{P}_{v, i}$ as centering around $\bar f(\mtx{Q}_i)$, 
we compare the variant Gating against Inducer-tuning.
While Gating has a weighted sum form similar to prefix-tuning, 
it suffers from a great performance drop on both tasks,
which justifies the effectiveness of our design for $\mtx{P}_{v, i}$'s.

\section{Conclusion}

In this work, we connect attention modules to Nadaraya-Watson kernel estimators and review prefix-tuning from a kernel estimator perspective.
We speculate that continuous prompt tuning prompts serve as inducing variables in kernel methods.
Following this principle, we propose inducer-tuning, which customizes an ``inducer'' for each query vector from the input sequence and adopts a residual form to resolve the initialization issue in prefix-tuning.
In addition, the perspective implies a potential limitation of prompt-based methods: the positional interactions in attention cannot adapt to the new tasks in prompt tuning.
We empirically demonstrate that our proposed method performs better on NLU and NLG tasks.

\section*{Limitations}

In this section, we start with a common limitation of the current parameter-efficient techniques, 
and further, discuss the specific limitation of our methods.

The shared limitation of parameter-efficient techniques is that they are not computation-efficient;
These methods choose to directly inherit the pre-trained weights of the backbone model and add some extra modules, which increases
the computational cost of these methods during inference.

Additionally, our method, which relies on the unique structure of self-attention, is only applicable to attention modules. And thus, not as generic as LoRA and adapters.

\section*{Ethics Statement}
As an efficient method for NLP, we consider our work to have a low ethical risk since the outcomes of the algorithm mainly depend on the downstream applications.
The usage of the method would be the same as some previous methods, i.e., the practical deployment for some applications.
Also, our method doesn't assume any specific structure of the input and thus doesn't leverage biases in the data.
We conclude that our work will not likely have a negative ethical impact.

\section*{Acknowledgements}

We appreciate all the valuable feedback from the anonymous reviewers.

\clearpage
\bibliography{custom}
\bibliographystyle{acl_natbib}

\clearpage
\clearpage
\appendix
\section{Dataset Details}
\label{sec:appendix_dataset}

\begin{itemize}[leftmargin=*]

\item The Multi-Genre Natural Language Inference Corpus \citep[\textbf{MNLI}]{N18-1101} involves $433$k sentence pairs of premises and hypotheses, labeled with textual entailment annotations.
The premise sentences include ten distinct genres, 
and the classification can be performed on both the matched (in-domain) and mismatched (cross-domain) sections.
Concatenating premises and hypothesis as the inputs, we obtain the sequence lengths are on average $39.9$ and max $444$.
For the results reported in Table~\ref{tab:nlu}, we follow \citet{hu2021lora} and take mismatched accuracy as the metric.

\item The Stanford Sentiment Treebank \citep[\textbf{SST2}]{socher2013recursive} is a corpus of movie reviews and human annotations of their sentiment. 
This task is incorporated into the GLUE benchmark \citep{wang2019glue}, 
and the dataset split assigns $67k$ sentences to the training set and $0.9k$ to the dev set. 
In SST2, the sequence lengths are on average $13.3$ and max $66$, much shorter than in MNLI.
As specified in the GLUE benchmark, we test the accuracy metric on whether the sentiment of a review sentence is positive or negative.

    \item The instances in \textbf{WebNLG} dataset are the mapping set of RDF triples to text.
They are Data/Text pairs,  where the ``Data'' is in a format of (subject, property, object) triples.
For the train and the validation set, they involve nine categories which are extracted from DBpedia;
while in the test set, there are five extra unseen categories,
which can partially reflect the generalization of the adaptation methods.
The input sequences in the training set consist of 1 to 7 triples, and the lengths of most sequences are bounded by 50 (as each triple only includes three short phrases). 
The official evaluation script is used in our experiments, and we report BLEU \citep{papineni2002bleu}, METEOR, \citep{lavie2007meteor} and TER \citep{snover2006study} as the metrics.

    \item \textbf{CoQA} is a large-scale dataset, mainly for conversational question answering. 
    It collects more than 8K conversations over text passages, involving over 127K questions with answers in 5 domains.
The average conversation length is 15 turns (each turn consists of a question and an answer).
The task requires the language model to generate answers to the given questions based on related conversation histories and documents in the dataset.
The average passage length in CoQA is 271 \citep[Table 3]{reddy2019coqa}.
We simply follow the evaluation script provided on the official website, reporting both the macro-average F1 score of word overlap and the exact-match metric~\citep{reddy2019coqa}.
\end{itemize}

\section{Training Details}
\label{sec:appendix_training}

\begin{table*}[h!]
\centering
\vskip 0.15in
\resizebox{\textwidth}{!}{
\begin{tabular}{|l|c|c|}
\hline
\textbf{Datasets}        & \textbf{Special tokens}       & \textbf{\# of trainable parameters for task embedding} \\
\hline
WebNLG          &
\begin{tabular}{@{}c@{}}
\texttt{<bos\_webnlg>}, \texttt{<eos\_webnlg>}, \texttt{<subject>}, \\
\texttt{<property>}, \texttt{<object>}, \texttt{<target\_webnlg>}
\end{tabular}
                                        & $6 * 768 = 4608$          \\ \hline
CoQA            &
\begin{tabular}{@{}c@{}}
\texttt{<bos\_qa>}, \texttt{<eos\_qa>}, \texttt{<question>}, \\
\texttt{<answer>}, \texttt{<document>}
\end{tabular}
                                        & $5 * 768 = 3840$    \\ \hline
\end{tabular}
}
\caption{\label{tab:tokens} The special tokens used in different tasks and the corresponding size of trainable parameters.}
\end{table*}

We mainly implement our methods based on the GitHub repositories provided by \citet{lin2020exploring} and \citet{he2021towards}.
Our code will be made public after the review procedure.

\subsection{General Training Settings}
\label{sec:task_embedding}

For the NLU tasks, we exactly follow the experimental setup used by \citet{he2021towards}, and more details can be found in Appendix~\ref{sec:hyperpara}.

For the two NLG tasks, we mainly follow the experimental setting adopted by \citet{lin2020exploring}, and specifically, keep using ``task embeddings'' in our experiments,
as they are also applied in the original GPT-2 model. 
These task embeddings are specialized segment embeddings used to indicate the different components of the text input (e.g., the three components of a triple in WebNLG, questions, and answers in CoQA, etc.).
\footnote{The task embedding for the special tokens will also be updated during training, though we do not count them in~\Cref{tab:result}.} 

We list the task embedding used in each NLG task:
for CoQA, we follow the task embedding suggested by \citet{lin2020exploring};
for WebNLG, we simply use the special tokens to indicate the different components in the triples.
The details of the special tokens in each task are summarized in Table~\ref{tab:tokens}.
Notably, the parameter budget for task embedding is much smaller than the number of tunable parameters in the aforementioned parameter-efficient adaptation methods (around 2M).

\subsection{Hyper-parameters for Training}
\label{sec:hyperpara}

For NLU tasks, we train the models with Adam \citep{kingma2015adam} optimizer and use a polynomial learning rate scheduler to make the learning rate linearly decay;
specifically, the learning rate is linearly warmed up from 0 for the first 6%
For NLG tasks, an AdamW \citep{loshchilov2018decoupled} optimizer is applied to train the models, and a linear learning rate scheduler with a 500-step warmup duration is used. 

For the evaluation of NLG tasks, we follow the script provided by \citet{lin2020exploring} to generate the texts through a greedy search for both WebNLG and CoQA.
As for the number of epochs and the argument for weight decay, we mainly follow the setting used by \citet{lin2020exploring, hu2021lora}:
for WebNLG, we train the model for 10 epochs; 
for CoQA, we train the model for 5 epochs.

For the model-specific hyper-parameters, namely batch size (gradient accumulation is used if necessary) and learning rate, 
we decide them for different methods based on the loss on the validation set.
For the proposed method inducer-tuning with/without LoRA and MAM inducer-tuning in Table~\ref{tab:result}, 
we set the learning rate as $0.00125$, and the batch size is $16$ for WebNLG;
for CoQA, the learning rate we use is $0.001$, and the batch size is $16$.
On MNLI, we set the learning rate as $0.0002$ for both two methods and the batch size as $32$ for inducer-tuning with LoRA and $16$ for MAM inducer-tuning;
on SST2, the learning rate is similarly set as $0.0002$, the batch size for inducer-tuning with LoRA is $16$, and for MAM inducer-tuning $64$.

To reduce the random variability in the results, all the methods reported are trained for multiple independent runs. 
In particular, for WebNLG, we train models over $5$ runs, and for CoQA, MNLI, and SST2 $3$ runs. 
The reported numbers in the cells in Tables~\ref{tab:result} and \ref{tab:nlu} are the mean value averaged over the runs,
and the significance tests in Table~\ref{tab:result} are also based on the replicates.

\subsection{Training Efficiency and Implementation Details}
\label{sec:implementation}

All the models in this work are implemented by PyTorch.
For the training runtime, if we perform the training using 1 Tesla V100 16GB GPU,
on WebNLG, it will take inducer-tuning and its variants around $8$ minutes to finish one epoch;
on CoQA, the time cost is around $160$ minute/epoch.
On MNLI and SST2, the runtime is $120$ and $25$ minute/epoch, respectively.
We remark all the parameter-efficient tuning methods have a similar time cost,
while indeed, they solely slightly save training time compared to fine-tuning.
The same phenomenon is also observed by \citet{lin2020exploring, ruckle2020adapterdrop}.

For the implementation of $\overline{\text{MLP}}_v$, we provide the exact expression for $\overline{\text{MLP}}_v^{(h)}(\mtx{Q}^{(h)})$ in head $h$ as follows:
\begin{align}
\sigma \left( \mtx{Q}^{(h)} \mtx{W}_1^{(h)} + \mtx{1} (\mtx{b}_1^{(h)})^T \right) \mtx{W}_2^{(h)} + \mtx{1} \mtx{b}_2^T,
\label{eqn:mlpv}
\end{align}
where $\sigma$ is the activation function.
As the superscript suggests,
$\mtx{W}_1^{(h)} \in \mb R^{p \times r}$, $\mtx{b}_1^{(h)} \in \mb R^{r}$, and $\mtx{W}_2^{(h)} \in \mb R^{r \times N_h p}$ are specific to the head $h$, 
while $\mtx{b}_2 \in \mb R^{N_h p}$ are shared among all the heads, which is the same case as in Equation~(\ref{eqn:wo}) (in the original attention sub-layer, the bias term $\mtx{b}_o$ applies to all the heads as well).

\subsection{Specific settings for baseline methods}
\label{sec:method_setting}

In this subsection, we provide the detailed setting for the methods in Tables~\ref{tab:result}, \ref{tab:nlu}, and \ref{tab:ablation} that need further specification.

In Table~\ref{tab:result}, the settings for Adapter-108 and Prefix-tuning-108 are clear, as the only arguments are the bottleneck size / prefix length;
for LoRA-54, we apply rank-54 updates for both $\mtx{W}_q$ and $\mtx{W}_v$, as suggested by \citet{hu2021lora};
for MAM adapter, we mimic the parameter assignment scheme (bottleneck size $512$ for FFN and prefix length $30$) by \citet{he2021towards}, and use the ratio $102:6$ to implement MAM adapters with $1.61$\% tunable parameters.

For the variants of inducer tuning, their settings are summarized in Table~\ref{tab:inducer_setting}.
In this table, the numbers in column $\text{MLP}_k$ and $\text{MLP}_v$ are the bottleneck sizes used for computing $\mtx{P}_{k}$ and $\mtx{P}_{v}$;
notice for Adaptive, the scope of virtual value tokens is not extended and thus has a larger bottleneck size than others. 
(Recall for $\overline{\text{MLP}}_v$, the size of $\mtx{W}_2^{(h)}$ in Equation~(\ref{eqn:mlpv}) is larger than the counterparts in $\text{MLP}_v$.
For the numbers in column \textit{LoRA}, they are the rank of the update used in the LoRA component to adjust $\mtx{W}_q$;
only for our proposed method inducer-tuning with LoRA, the number will be non-zero.

\begin{table*}[h!]
\centering
\vskip 0.15in
\begin{tabular}{|l|cccc|}
\hline
NLG tasks  & $\text{MLP}_k$ & $\text{MLP}_v$ & LoRA & FFN-adapter\\
\hline
Adaptive       & 0      & 108    & 0    & 0\\
Extension      & 0      & 16     & 0    & 0\\
Gating         & 10     & 15     & 0    & 0\\
\hline
Inducer-tuning & 10     & 15     & 0    & 0\\
$\sim$ w/ LoRA & 5      & 12     & 24   & 0\\ 
MAM inducer-tuning & 3     & 7     & 16    & 42\\
\hline
\hline
NLU tasks  & $\text{MLP}_k$ & $\text{MLP}_v$ & LoRA & FFN-adapter\\
\hline
Inducer-tuning & 6 & 4 & 0 & 0 \\
Inducer-tuning + LoRA & 2      & 4     & 4   & 0 \\ 
MAM inducer-tuning & 2     & 2     & 4    & 12 \\
\hline
\end{tabular}
\caption{\label{tab:inducer_setting} The exact parameter assignment settings for variants of inducer-tuning in Tables~\ref{tab:result}, \ref{tab:nlu}, and \ref{tab:ablation}. 
}
\end{table*}

\section{Attention as Kernels}
\label{sec:kernel_estimator}

To justify the claim that attention is a kernel operation, we construct a Nadaraya–Watson kernel estimator \citep[Definition~5.39]{wasserman2006all} of a query vector $\mtx{Q}_i$ (taking $\{\mtx{K}_j\}_{j=1}^n$ as the supporting points) as follows:
\begin{align}
\label{eq:kernel_estimator}
& f(\mtx{Q}_i) = \sum_{j=1}^n \ell_j(\mtx{Q}_i) \mtx{C}_j, \\
& \text{where} \quad \ell_j(\mtx{Q}_i) \defeq \frac{\kappa(\mtx{Q}_i, \mtx{K}_j)}{\sum_{k=1}^n \kappa(\mtx{Q}_i, \mtx{K}_j)}. \nonumber
\end{align}
$\kappa(\cdot, \cdot)$ is a kernel function, 
and $\mtx{C}_j$'s are the coefficients corresponding to the rows $\mtx{V}_j$'s in the value matrix $\mtx V$.

Take kernel function $\kappa(x, y) = \exp\left(\dotp{x}{y} / \sqrt{p} \right)$.
We slightly abuse the notation $\kappa(\mtx{Q}, \mtx{K})$ to represent the $n$-by-$n$ empirical kernel matrix $\mtx{M}$, 
in which the $i$-th row and the $j$-th column is $\kappa(\mtx{Q}_i, \mtx{K}_j), \forall i \in [n], j \in [N]$.
With these notations, the output of the kernel estimator will be,
\begin{align}
\label{eq:matrix_form}
\mtx{D}^{-1} \mtx{M} \mtx{C},
\end{align}
where $\mtx{D}$ is a diagonal matrix serving as the row normalization in Equation~(\ref{eq:kernel_estimator}), and $\mtx{C}$ is an $n$-by-$p$ matrix with $\mtx{C}_j$ as its $j$-th row.
We observe an obvious correspondence between Equation~(\ref{eq:matrix_form}) and the standard attention in Equation~(\ref{eqn:attn}).
The correspondence implies a finer division of the attention module: the empirical kernel matrix $\mtx{M}$ ($\mtx{D}$ is decided by $\kappa(\mtx{Q}, \mtx{K})$) and the value part $\mtx{C}$. 
(In Section~\ref{sec:extension}, we show that $\mtx{C}$ includes but is \textit{not} limited to the \textit{value} matrix in attention.)

\section{Example}

We provide an example answer generated by fine-tuning and our inducer-tuning with LoRA on CoQA in Table~\ref{tab:example}.

\begin{table*}[!htbp]
    \label{tab:addlabel}%
    \centering
    \vskip 0.15in
    \begin{tabularx}{\textwidth}{|l|X|}
        \hline
        \textbf{Documents} &  
        Kendra and Quinton travel to and from school every day. Kendra lives further from the bus stop than Quinton does, stops every morning at Quinton's house to join him to walk to the bus stop. Every afternoon, after school, when walking home from the bus stop they go in for cookies and milk that Quinton's mother has ready and waiting for them. Quinton can't eat cheese or cake so they had the same snack every day. They both work together on their homework and when they are done they play together. Kendra always makes sure to leave in time to get home for dinner. She doesn't want to miss story time which was right before bedtime. \newline \newline
        One morning Kendra walked up to Quinton's house, \textbf{she thought something might be wrong} because normally Quinton was waiting outside for her and on this morning \textbf{he was not to be found}. Kendra went up to the door and knocked. She waited and waited and yet \textbf{no one answered}. She saw that Quinton's mother's car wasn't in their driveway which was weird. She waited for a few bit looking up and down the block and getting worried when Quinton was nowhere to be found. \newline \newline
        Kendra didn't want to miss the bus to school and hurried off to make it in time. The bus driver saw that she was upset and that Quinton was not with her that morning. She told him what happened and he said that he was sure that everything would be okay. \newline \newline
        Kendra got to school, ran to her teacher and told him what happened that morning. The teacher smiled and told her not to worry, Quinton's mother had called and he was going to the dentist and would be at school after lunch and that she would see him at the bus stop like normal tomorrow. \newline \newline
        Q4: What happened when Kendra knocked on Quinton's door? \newline 
        Reference: no one answered \\
        \hline
        Fine-tuning &
        she thought something might be wrong \\
        \hline
        Inducer-tuning w/ LoRA &
        he was not to be found \\
        \hline
    \end{tabularx}%
\caption{\label{tab:example} A qualitative example from CoQA. In particular fine-tuning generates an answer farther away from the correct answer than inducer-tuning with LoRA.}
\end{table*}%

\end{document}